%% file: PL_Main.tex
\documentclass[10pt,twocolumn,letterpaper]{article}

\usepackage{cvpr}
\usepackage{times}
\usepackage{epsfig}
\usepackage{graphicx}
\usepackage{amsmath}
\usepackage{amssymb}
\usepackage{enumitem}

\usepackage{flexisym}
\usepackage{ifthen}
\usepackage{color}
\usepackage{bigstrut}
\usepackage{multirow}
\usepackage{appendix}

\usepackage[pagebackref=true,breaklinks=true,letterpaper=true,colorlinks,bookmarks=false]{hyperref}

\graphicspath{{figures/}{./}}

\definecolor{frenchblue}{rgb}{0.0, 0.45, 0.73}
\definecolor{gray}{rgb}{0.5,0.5,0.5} 
\definecolor{green}{rgb}{0, 0.4, 0} 
\definecolor{orange}{rgb}{1, 0.5, 0} 	
\definecolor{mahogany}{rgb}{0.75, 0.25, 0.0}
\definecolor{purple}{rgb}{0.6, 0, 0.6}
\definecolor{darkgreen}{rgb}{0, 0.4, 0.4} 
\definecolor{blue}{rgb}{0.0, 0.45, 0.73}
\definecolor{aaaa}{rgb}{0.55, 0.1, 0.7}
\definecolor{red}{rgb}{1.0, 0, 0}

\newboolean{revising}
\setboolean{revising}{false}
\ifthenelse{\boolean{revising}}
{
	\newcommand{\fuen}[1]{\textcolor{blue}{[FuEn]: #1}}
	\newcommand{\sunda}[1]{\textcolor{blue}{[SunDa]: #1}}
	\newcommand{\james}[1]{\textcolor{blue}{[James]: #1}}
	\newcommand{\markred}[1]{\textcolor{red}{#1}}

} {
	\newcommand{\fuen}[1]{}
	\newcommand{\sunda}[1]{}
	\newcommand{\james}[1]{}
	\newcommand{\markred}[1]{#1}

}

\newcommand{\tb}[1]{\textbf{#1}}

\cvprfinalcopy 

\def\cvprPaperID{1441} 

\ifcvprfinal\fi
\begin{document}

\title{DuLa-Net: A Dual-Projection Network for Estimating Room Layouts from a Single RGB Panorama}

\author{
Shang-Ta Yang$^{1,2}$\\
{\tt\small sundadenny@gapp.nthu.edu.tw}
\and
Fu-En Wang$^1$\\
{\tt\small fulton84717@gapp.nthu.edu.tw}
\and
Chi-Han Peng$^2$\\
{\tt\small pchihan@asu.edu}
\and
Peter Wonka$^2$\\
{\tt\small pwonka@gmail.com}
\and
Min Sun$^1$\\
{\tt\small sunmin@ee.nthu.edu.tw}
\and
Hung-Kuo Chu$^1$\\
{\tt\small hkchu@cs.nthu.edu.tw}
\and
$^1$National Tsing Hua University\\
\and
$^2$KAUST\\
}

\maketitle

\input{PL_Macros.tex}
\input{PL_Abstract.tex}



\input{PL_Introduction.tex}

\input{PL_Related.tex}

\input{PL_Overview.tex}

\input{PL_E2P.tex}

\input{PL_Network.tex}

\input{PL_LayoutFitting.tex}

\input{PL_Dataset.tex}

\input{PL_Experiments.tex}

\input{PL_Conclusion.tex}


{\small
\bibliographystyle{ieee_fullname}
\bibliography{PL_bib}
}

\begin{appendices}
\input{PL_Supplementary.tex}

\end{appendices}

\end{document}

%% file: PL_Macros.tex

\renewcommand{\etal}{\textit{et~al.}}
\newcommand{\fref}[1]{{Fig.~\ref{#1}}}
\newcommand{\tref}[1]{{Table~\ref{#1}}}
\newcommand{\sref}[1]{{Sec.~\ref{#1}}}
\newcommand{\eref}[1]{{Eqn.~\ref{#1}}}

\newcommand{\networkName}{DuLa-Net} 
\newcommand{\EtoP}{E2P}
\newcommand{\ceilingbranch}{$B_C$ }
\newcommand{\ceilingencoder}{$E_{B_C}$ }
\newcommand{\ceilingdecoder}{$D_{B_C}$ }
\newcommand{\panoramabranch}{$B_P$ }
\newcommand{\panoramaencoder}{$E_{B_P}$ }
\newcommand{\panoramadecoder}{$D_{B_P}$ }
\newcommand{\fcmap}{M_{FC}}
\newcommand{\fpmap}{M_{FP}}
\newcommand{\fcName}{floor-ceiling probability map}
\newcommand{\fpName}{floor plan probability map}
\newcommand{\ffpName}{fused floor plan probability map}
\newcommand{\fcmapCeil}{\fcmap^{C}}
\newcommand{\fcmapFloor}{\fcmap^{F}}
\newcommand{\ceilView}{ceiling-view}
\newcommand{\panoView}{panorama-view}
\newcommand{\ceilBranch}{ceiling-branch}
\newcommand{\panoBranch}{panorama-branch}
\newcommand{\cameraHeight}{$H_{ceiling}$}

\newcommand{\dsName}{Realtor360} 
\newcommand{\dsSize}{2573}  
\newcommand{\sunSize}{593}  
\newcommand{\ycSize}{1980}  
\newcommand{\tnSize}{2169}  
\newcommand{\ttSize}{404}  

\newcommand{\metricND}[1]{\footnotesize{#1D IoU (\%)}}

%% file: PL_Abstract.tex

\begin{abstract}

We present a deep learning framework, called {\networkName}, to predict Manhattan-world 3D room layouts from a single RGB panorama.
To achieve better prediction accuracy, our method leverages two projections of the panorama at once, namely the equirectangular {\panoView} and the perspective {\ceilView}, that each contains different clues about the room layouts.
Our network architecture consists of two encoder-decoder branches for analyzing each of the two views. In addition, a novel feature fusion structure is proposed to connect the two branches, which are then jointly trained to predict the 2D floor plans and layout heights.
To learn more complex room layouts, we introduce the {\dsName} dataset that contains panoramas of Manhattan-world room layouts with different numbers of corners.
Experimental results show that our work outperforms recent state-of-the-art in prediction accuracy and performance, especially in the rooms with non-cuboid layouts.
\end{abstract}

%% file: PL_Introduction.tex

\section{Introduction}
\label{sec:intro}
Inferring high-quality 3D room layouts from indoor panoramic images plays a crucial role in indoor scene understanding and can be beneficial to various applications, including virtual/augmented reality and robotics.
To that end, recent methods recover 3D room layouts by using deep learning to predict the room corners and boundaries on the input panorama. For example, LayoutNet~\cite{layoutnet} achieved impressive reconstruction accuracy for Manhattan world-constrained rooms.
However, the clutter in the room, e.g. furniture, poses a challenge to extract critical corners and edges that are occluded in the input panorama.
In addition, estimating 3D layouts from 2D corner and edge maps is an ill-posed problem and thus imposing extra constraints in the optimization. 
Therefore, it remains challenging to process complex room layouts.
%

\input{./figures/PL_fig_teaser.tex}

In this work, we present a novel end-to-end framework to estimate a 3D room layout from a single RGB panorama.
By the intuition that a neural network may extract different kinds of features given the same panorama but in different projections, we propose to predict the room layouts from two distinct views of the panoramas, namely the equirectangular \emph{\panoView} and the perspective \emph{\ceilView}.
%
%
The network architecture follows the encoder-decoder scheme and consists of two branches, the \emph{\panoBranch} and the \emph{\ceilBranch}, for respectively analyzing images of the {\panoView} and the {\ceilView}.
The outputs of {\panoBranch} include a \emph{\fcName} and a \emph{layout height}, while the {\ceilBranch} outputs a \emph{\fpName}.
To share information between branches, we employ a feature fusion scheme to connect the first few layers of decoders through a \emph{{\EtoP}} conversion that transforms intermediate feature maps from equirectangular projection to perspective {\ceilView}.
We find that better prediction performance is achieved by jointly training the two connected branches.   
The final 2D floor plan is then obtained by fitting an axis-aligned polygon to a \emph{\ffpName} (see Figure~\ref{fig:floorplan_fitting} for details) and then extruded by the estimated layout height.

To learn from panoramas with complex layouts, we need a proper dataset for network training and testing.
However, existing public datasets, such as PanoContext~\cite{panocontext} dataset, provide mostly labeled 3D layouts with simple cuboid shapes.
To learn more complex layouts, we introduce a new dataset, \emph{\dsName}, which includes a subset of SUN360~\cite{sun360} dataset (\markred{593} living rooms and bedrooms) and \markred{1980} panoramas collected from a real estate database.
We annotated the whole dataset with a custom-made interactive tool to obtain the ground-truth 3D layouts.

A key feature of our dataset is that it contains rooms with more complex shapes in terms of the numbers of the corners.
The experimental results demonstrate that our method outperforms the current state-of-the-art method (\cite{layoutnet}) in prediction accuracy, especially with rooms with more than four corners. Our method also takes much less time to compute the final room layouts.
\fref{fig:teaser} shows some room layouts estimated by our method.
Our contributions are summarized as follows:
\begin{itemize}
    \item We propose a novel network architecture that contains two encoder-decoder branches to analyze the input panorama in two different projections. These two branches are further connected through a {\em feature fusion} scheme. This {\em dual-projection} architecture can infer room layouts with more complex shapes beyond cuboids and L-shapes.
    \item Our neural network is an important step towards building an {\em end-to-end} architecture. Our network directly outputs a probability map of the 2D floor plan. This output requires significantly less post-processing to obtain the final 3D room layout than the output of the current state of the art.
    %
    %
    \item We introduce a new data set, called {\dsName}, that contains {\dsSize} panoramas depicting rooms with 4 to 12 corners. To the best of our knowledge, this is largest data set of indoor images with room layout annotations currently available.
\end{itemize}

\if 1
Inferring high-quality room layouts from indoor panorama images can be beneficial to several applications in computer vision and computer graphics, including indoor scene understanding, VR, AR, and robotics.  The challenge of these methods is to ignore the clutter in the room, e.g. furniture, and extract the layout even though critical corners and edges can be occluded in the input images. In our work we would like to improve on very exciting recent deep learning frameworks~\cite{} in two ways. First, we would like to be able to detect more complex room shapes than just cuboid and L-shaped rooms. Second, we improve the overall quality of the reconstruction even for simpler rooms. 

In this paper, we present our deep learning architecture Pano2Layout to estimate complex room 3D layouts from a single color panorama. Our reconstruction concept is decomposing the layout estimation into the floor plan and layout height prediction. Instead of extracting the 2D features on equirectangular, we tend to directly predict the floor plan in another projection.  We found that in the different projections of the panorama, each has the advantage to extract the floor plane.  To leverage this property, we design a deep learning model with two branches in two projection (equirectangular and up-view). 

First, our input is a Manhattan-aligned panorama. We use the same strategy as PanoContext to analyzes the vanishing points and make sure the panorama align the Manhattan World. Then our deep learning model joint predicts the floor plane probability map, floor-ceiling probability map, and layout height. For more detail,  our CNN model including two branches: equirectangular and up-view branch. Each of them using the encode-decode structure to predict floor plane map and floor-ceiling map separately. We also design a warping module to fuse the feature in two branches The experiments show that this fusion can overall improve the floor plane prediction. At the Last step, our method using a simple optimation to fit the 2D floor plan geometry on floor plane probability map, and using the layout height estimation result to extend to 3D room layout.

Our Method can infer simpler room layouts with 4(cuboid) and 6(L-shape) as well as complex layouts with up to 12 corners. Current publication panorama dataset is lack of complex room label, so we collect our new dataset from the estate company and design an annotator to do careful layout labeling. We also relabel the SUN360 for more detail. 

\begin{itemize}
    \item We provide a new network architecture that can estimate more complex indoor layouts from panorama images than the current state of the art.
    \item We can improve the accuracy of the current state of the art on simpler layouts.
    \item We introduce an interactive tool that we used to create a new dataset Complex360 with XXXX annotated panoramas.
\end{itemize}
\fi

%% file: figures/PL_fig_teaser.tex

\begin{figure}
\begin{center}
  \includegraphics[width=\columnwidth]{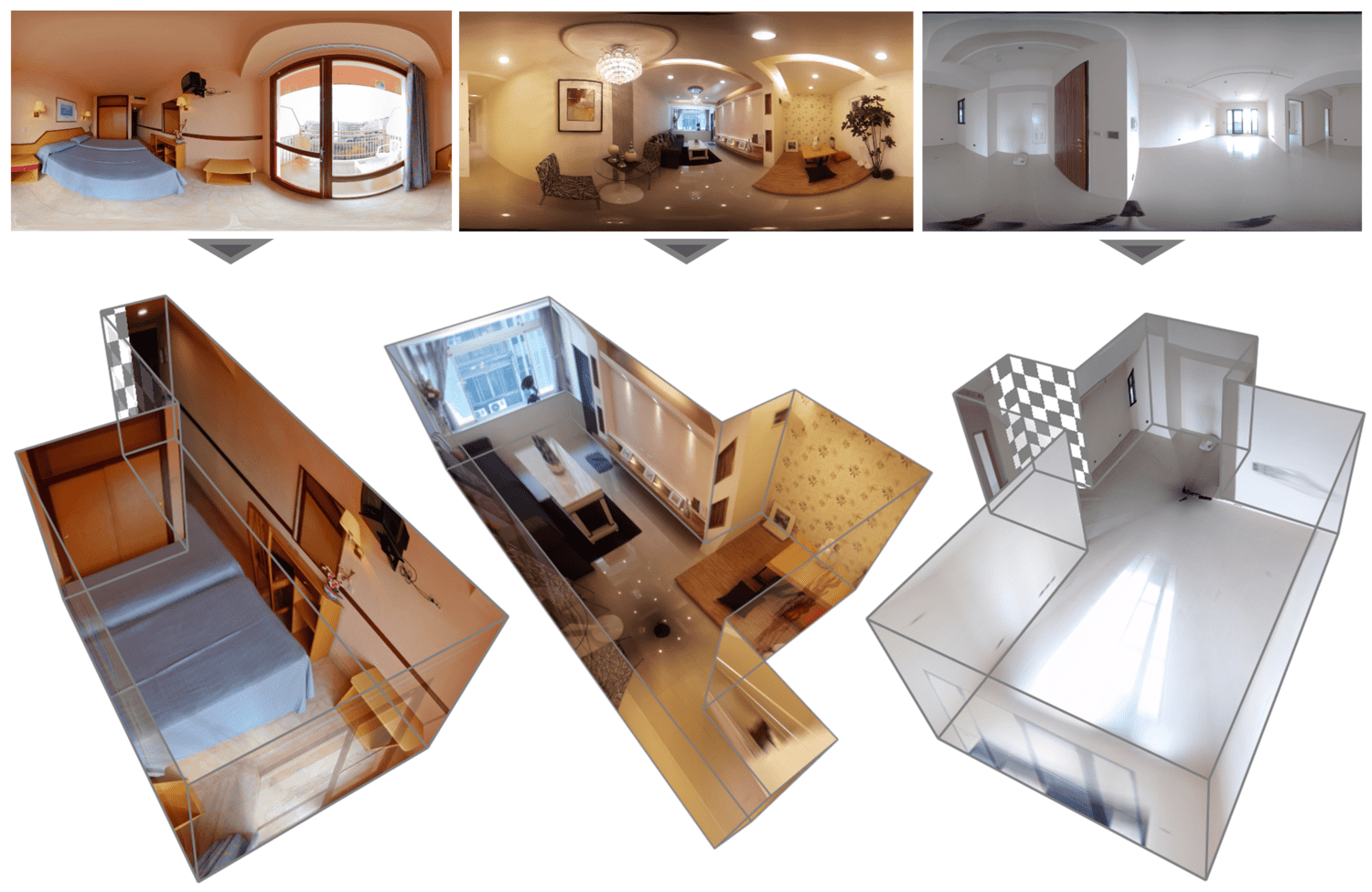}
  \caption{3D room layouts with different complexity are estimated from a single RGB panorama using our system. (\emph{Left to right}) Room layout with a floor plan of 6 corners, 8 corners, and 10 corners. The checkerboard patterns on the walls indicate the missing textures due to occlusion.}
  \label{fig:teaser}
\end{center}
\end{figure}

%% file: PL_Related.tex

\input{./figures/PL_fig_overview.tex}

\section{Related Work}

\noindent

There are multiple papers that propose a solution to estimate room layouts from a single image taken in an indoor environment. They mainly differ in three aspects: 1) the assumptions of the room layouts, 2) the types of the input images, and 3) the methods. In terms of room layout assumptions, a popular choice is the "Manhattan world" assumption~\cite{manhattan}, meaning that all walls are aligned with a global coordinate system~\cite{manhattan,mjc}. To make the problem easier to solve, a more restrictive assumption is that the room is a cuboid~\cite{5459411, delay, roomnet}, i.e., there exist exactly four room corners. Our method adopts the Manhattan world assumption but allows for arbitrary numbers of corners. 

In terms of types of input images, the images may differ in the FoV (field of view) - ranging from being monocular (i.e., taken from a standard camera) to 360$^\circ$ panoramas, and whether depth information is provided. The methods then largely depend on the input image types. The problem is probably most difficult to solve when only a monocular RGB image is given. Typically, geometric (e.g., lines and corners)~\cite{5206872, 5459411, Ramalingam2013Lifting3M} and/or semantic (e.g., segmentation into different regions~\cite{1541316, Hoiem2007} and volumetric reasoning~\cite{NIPS2010_4120}) "cues" are extracted from the input image, a set of room layout hypotheses is generated, and then an optimization or voting process is taken to rank and select one among the hypotheses. Recently, neural network-based methods took stride in tackling this problem. A trend is that the neural networks generate higher and higher levels of information - starting from line segments~\cite{informative, stpio}, surface labels~\cite{delay}, to room types~\cite{roomnet} and room boundaries and corners~\cite{layoutnet}, to make the final layout generation process increasingly easier to solve. Our method pushes this trend one step further by using neural networks to directly predict a 2D {\fpName} that requires only a 2D polygon fitting process to produce the final 2D room layout.

If depth information is provided, there exist methods that estimate scene annotations including room layouts~\cite{6751268, 7780394, deepcontext}. A deeper discussion is beyond the scope of this paper.

Closely related problems include depth estimation from a given image~\cite{omnidepth, ceilingview} and scene reconstructions from point clouds~\cite{6162880, Monszpart:2015:RRM:2809654.2766995, FloorNet}. Note that neither estimated depths nor reconstructed 3D scenes necessarily equate a clean room layout as such inputs may contain clutters.

\noindent\textbf{360$^\circ$ panorama:} The seminal work by Zhang et al.~\cite{panocontext} advocates the use of 360$^\circ$ panoramas for indoor scene understanding for the reason that the FOV of 360$^\circ$ panoramas is much more expansive. Work in this direction flourished, including methods based on optimization approaches over geometric~\cite{Fukano2016RoomRF,ceilingview,efficient3d} and/or semantic cues~\cite{7926629,automatic} and later based on neural networks~\cite{roomnet,layoutnet}. Except for LayoutNet~\cite{layoutnet}, most methods rely on leveraging existing techniques for single perspective images on samples taken from the input panorama. We believe that this is a major reason of LayoutNet's superior performance since it performs predictions on the panorama as a whole, thus extracting more global information that the input panorama might contain. A further step in this direction can be found in~\cite{ceilingview}, in which the input panorama is projected to a 2D "floor" view in which the camera position is mapped to the center of the image and the vertical lines in the panorama become radial lines emanated from the image center. An advantage of this approach is that the room layout becomes a 2D closed loop that can be extracted more easily. We derived our "ceiling" view idea here - instead of looking {\em downward} toward the floor in which all the clutter in the room is included, we look {\em upward} toward the ceiling and got a more clutter-free view of the room layout.

%% file: figures/PL_fig_overview.tex

\begin{figure*}[!t]
  \centering
  \includegraphics[width=\textwidth]{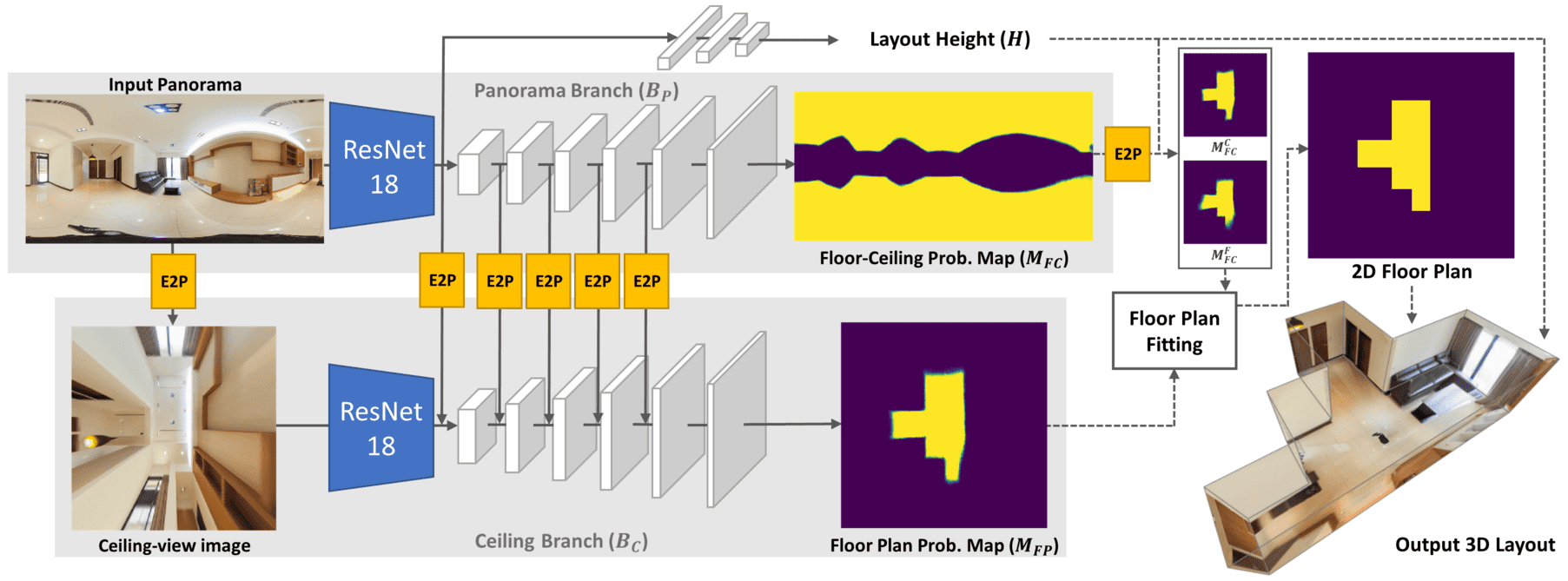}
  \caption{
  Our network architecture follows the encoder-decoder scheme and consists of two branches. Given a panorama in equirectangular projection, we additionally create a perspective {\em ceiling-view} image through a equirectangular-to-perspective ({\EtoP}) conversion. The panorama and the ceiling-view images are then fed to the {\em panorama-view} (upper) and {\em ceiling-view} (lower) branches. A {\EtoP}-based feature fusion scheme is employed to connect the two branches, which are jointly trained by the network to predict: 1) probability maps of the floor and ceiling in panorama view, 2) a floor plan in ceiling view, and 3) a layout height. Then, our system estimates a 2D floor plan by fitting a Manhattan-world aligned polygon to a weighted average of the three floor plans, which is further extruded using the predicted layout height to obtain the final 3D room layout.
  }
  \label{fig:overview}
\end{figure*}

%% file: PL_Overview.tex
\section{Overview}
\label{sec:overview}
\fref{fig:overview} illustrates the overview of our framework.
Given the input as an equirectangular panoramic image, we follow the same pre-processing step used in PanoContext~\cite{panocontext} to align the panoramic image with a global coordinate system, i.e. we make a Manhattan world assumption.
Then, we transform the panoramic image into a perspective {\ceilView} image through an equirectangular to perspective (\EtoP) conversion (\sref{sec:e2p}).
The {\panoView} and {\ceilView} images are then fed to a network consisting of two encoder-decoder branches. These two branches are connected via a {\EtoP}-based feature fusion scheme and jointly trained to predict a {\fpName}, a {\fcName}, and the layout height (\sref{sec:network}).
%
%
%
Two intermediate probability maps are derived from the {\fcName} using {\EtoP} conversion and combined with {\fpName} to obtain a {\ffpName}.
The final 3D Manhattan layout is determined by extruding a 2D Manhattan floor plan estimated on the {\ffpName} using the predicted layout height (\sref{sec:layoutFitting}).

%% file: PL_E2P.tex


\section{{\EtoP} conversion}
\label{sec:e2p}
In this section, we explain the formulation of {\EtoP} conversion that transforms an equirectangular panorama to a perspective image. We assume the perspective image is square with dimension $w \times w$.
For every pixel in the perspective image at position $(p_x,p_y)$, we derive the position of the corresponding pixel in the equirectangular panorama, $(p\textprime_x,p\textprime_y)$, $-1 \le p\textprime_x \le 1, -1 \le p\textprime_y \le 1$, as follows.
First, we define the field of view of the pinhole camera of the perspective image as $FoV$. Then, the focal length can be derived as:
$$
f = 0.5 * w * \cot(0.5 * \mathrm{FoV})~.
$$

$(p_x, p_y, f)$, the 3D position of the pixel in the perspective image in the camera space, is then rotated by 90$^\circ$ or -90$^\circ$ along the x-axis (counter-clockwise) if the camera is looking upward (looking at the ceiling) or downward (looking at the floor), respectively.

Next, we project the rotated 3D position to the equirectangular space. To do so, we first project it onto a unit sphere by vector normalization, $(s_x, s_y, s_z)$, and apply the following formula:
\begin{align}
(p\textprime_{x}, p\textprime_{y}) &= (\frac{arctan_2 (\frac{s_{x}}{s_{z}})}{\pi}~, \frac{arcsin (s_{y})}{0.5\pi}),
\end{align}
\noindent to project $(s_x, s_y, s_z)$, the 3D position on the unit sphere, back to $(p\textprime_x,p\textprime_y)$, the corresponding 2D position in the equirectangular panorama. Finally, we use $(p\textprime_x,p\textprime_y)$ to interpolate a pixel value from the panorama. We note that this process is differentiable so it can be used in conjunction with backpropagation.

%
%
%
%
%
%
%
%
%

\if 1
\fuen{
For an equirectangular panorama, we can treat its image plane as a sphere. Given the image dimension, viewing direction and FoV, we can recover the corresponding perspective image. To achieve this, we need to derive the image plane of the perspective image. With the given FoV and dimension, here we assume the dimension is $w \times w$, we can obtain the focal length of this plane as 
\begin{align}
f = 0.5 * w * \cot(0.5 * FoV)~.
\end{align}
then we can derive the 3D grid $ G $ of pixels in this planes as we do in pinhole camera. As shown in Fig.~\ref{fig:projection}, for any 3D point in the world coordinate $P_{xyz}$, the corresponding projected location of $P_{xyz}$ in perspective and equirectangular image is $p\textprime$ and $p$, respectively. For applying backward warping to obtain the color of perspective image, $p\textprime$ should be the points in the 3D grid $G$. Now we can assume the sphere is a unit sphere then $p = \frac{p\textprime}{|p\textprime|}$. According equirectangular projection, the mapping relationship from $p$ to the 2D pixel in equirectangular image will follow:
\begin{align}
    x &= \frac{arctan_2 (\frac{p_x}{p_z})}{\pi}~, \\
    y &= \frac{arcsin (p_y)}{0.5\pi}~.
\end{align}
where x, y is the normalized 2D location in equirectangular image, while $p_x$, $p_y$ and $p_z$ are the 3D location of $p$. The corresponding perspective image can be recovered by the interpolation according to the x, y. However, in addition to applying this technique in color image, we can also apply it to feature space of equirectangular image from deep neural network and this procedure is fully differentiable. In this paper, we use this technique to convert equirectangular image/feature into both ceiling and floor perspective image/feature with 160$^\circ$ FoV and we call this operation as \textbf{E2P (equirectangular to perspective)} module.
}

The usual approach for such image transformations is to perform the inverse mapping. That is, one needs to consider each pixel in the output image (perspective) and map backwards to find the best estimate pixel in the input image (spherical). In this way every pixel in the output image is found (compared to a forward mapping), it also means that the performance is governed by the resolution of the output image (and supersampling) irrespective of the size of the input image. A key aspect of these mappings is also to perform some sort of antialiasing, the solutions here use a simple supersampling approach.

The default usage creates an image with a 100 degree horizontal field of view, the vertical field of view is calculated based upon the image dimensions, the default being 1920x1080. Zero degrees longitude is 1/4 of the width of the image from the left edge.
\fi

%% file: PL_Network.tex

\section{Network architecture}
\label{sec:network}

Our network architecture is illustrated in \fref{fig:overview}. It consists of two encoder-decoder branches, for the panorama-view and the ceiling-view input images. We denote the panorama-view branch as \panoramabranch and the ceiling-view branch as \ceilingbranch. The encoder and decoder of \panoramabranch are denoted as \panoramaencoder and \panoramadecoder and for \ceilingbranch they are denoted as \ceilingencoder and \ceilingdecoder. A key concept is that our network predicts the floor plan and the layout height. With these two predictions, we can reconstruct a 3D room layout in a post-process (\sref{sec:layoutFitting}).

\subsection{Encoder}
\label{sec:encoder}

We use ResNet-18 as the architecture for both \panoramaencoder and \ceilingencoder. The input dimension of \panoramaencoder is $512 \times 1024 \times 3$ (the dimension of the input panorama) and the output dimension is $16 \times 32 \times 512$. For \ceilingencoder, the input and output dimensions are $512 \times 512 \times3$ and $16 \times 16 \times 512$. Note that the input of \ceilingencoder is a perspective ceiling-view image generated by applying E2P conversion to the input panorama with $FoV$ set to 160$^\circ$ and $w$ set to 512.
%
%
We also tried other more computationally expensive network architectures such as ResNet-50 for the encoders. However, we find no improvements in accuracy so we chose to work with ResNet-18 for simplicity.

\subsection{Decoder}
\label{sec:decoder}

Both \panoramadecoder and \ceilingdecoder consist of six convolutional layers. The first five layers are $3 \times 3$ resize convolutions~\cite{resizeconv} with ReLU activations. The last layer is a regular $3 \times 3$ convolution with sigmoid activation. The numbers of channels of the six layers are 256, 128, 64, 32, 16, and 1. To infer the layout height, we add three fully connected layers to the middlemost feature of \panoramabranch. The dimensions of the three layers are 256, 64, and 1. To make the regression of the layout height more robust, we add dropout layers after the first two layers. To take the middlemost feature as input, we first apply global average pooling along both x and y dimensions, which produces an 1-D feature with 512 dimension, and take it as the input of the fully connected layers. 

The output of \panoramabranch is a probability map of the floor and the ceiling in the equirectangular projection, denoted as the \emph{\fcName} ($M_{FC}$). For \ceilingbranch, the output is a probability map of the floor plan in the ceiling view, denoted as the \emph{\fpName} ($M_{FP}$). Note that \panoramabranch also outputs a predicted layout height ($H$).

\subsection{Feature fusion}
\label{sec:fusion}

We find that applying fusion techniques to merge the features in both \panoramabranch and \ceilingbranch increases the prediction accuracy. We conjecture a reason as follows. In a ceiling-view image, the areas near the image boundary (where some useful visual clues such as shadows and furniture arrangements exist) are more distorted, which can have a detrimental effect for the ceiling-view branch to infer room structures. By fusing features from the panorama-view branch (in which distortion is less severe), performance of the ceiling-view branch can be improved.




We apply fusions before each of the first five layers of {\panoramadecoder} and {\ceilingdecoder}. For each fusion connection, a E2P conversion (\sref{sec:e2p}) with $FoV$ set to 160$^\circ$ is taken to project the features in {\panoramadecoder}, which are originally in the equirectangular view, to the perspective ceiling view. Each fusion works as follows:
\begin{align}
\label{equ:fusion}
    f_{B_C}^* = f_{B_C} + \frac{\alpha}{\beta^i} \times f_{B_P},~i \in \{0, 1, 2, 3, 4\},
\end{align}
where $f_{B_C}$ is the feature from {\ceilingbranch} and $f_{B_P}$ is the feature from {\panoramabranch} after applying the {\EtoP} conversion. $\alpha$ and $\beta$ are the decay coefficients. $i$ is the index of the layer. After each fusion, the merged feature, $f_{B_C}^*$, is sent into the next layer of {\ceilingdecoder}. The performance improvement of this technique is discussed in \sref{sec:exp}. 

\subsection{Loss function}
\label{sec:loss}

For $M_{FC}$ and $M_{FP}$, we apply binary cross entropy loss:
\begin{align}
    E_b(x, x^*) = -\sum_{i} {x}_i^*\log({x}_i) + (1-{x}_i^*)\log(1-{x}_i).
\end{align}
For $H$ (layout height), we use L1-loss:
\begin{align}
    E_{L1}(x, x^*) = \sum_{i} |x_i - x^*_i|.
\end{align}
The overall loss function is:
\begin{align}
    \label{equ:loss}
    &L = E_b(M_{FC}, M_{FC}^*) + E_b(M_{FP}, M_{FP}^*) + \gamma E_{L1}(H, H^*),
\end{align}
where $M_{FC}^*$, $M_{FP}^*$ and $H^*$ are the ground truth of $M_{FC}$, $M_{FP}$, and $H$.

\subsection{Training details}
\label{sec:training}

We implement our method with PyTorch\cite{pytorch}. We use the Adam\cite{adam} optimizer with $\beta_1=0.9$ and $\beta_2=0.999$. The learning rate is $0.0003$ and batch size is $4$. Our training loss converges after about $120$ epochs. For each training iteration we augment the input panorama with random flipping and horizontal rotatations by 0$^\circ$, 90$^\circ$, 180$^\circ$, and 270$^\circ$. For fusion, we set $\alpha$ and $\beta$ in \eref{equ:fusion} to be $0.6$ and $3$. We set the $\gamma$ in \eref{equ:loss} to be $0.5$. 
\markred{
Because we estimate the {\fpName} in the ceiling view, we assume the distance between the camera and the ceiling to be 1.6 meters, and use this constant to normalize the ground truth.
}
%


%% file: PL_LayoutFitting.tex

\input{./figures/PL_fig_floorplan_fitting.tex}

\section{3D layout estimation}
\label{sec:layoutFitting}
Given the probability maps ($\fcmap$ and $\fpmap$) and the layout height ($H$) predicted by the network, we reconstruct the final 3D layout in the following two steps:
\begin{enumerate}
    \item Estimating a 2D Manhattan floor plan shape using the probability maps.
    \item Extruding the floor plan shape along its normal according to the layout height.
\end{enumerate}
%
%
For step 1, two intermediate maps, denoted as $\fcmapCeil$ and $\fcmapFloor$, are derived from ceiling pixels and floor pixels of the {\fcName} using the {\EtoP} conversion.
\markred{
We further use a scaling factor, $1.6/(H - 1.6)$, to register the $\fcmapFloor$ with $\fcmapCeil$, where the constant $1.6$ is the distance between  the camera and the ceiling.
}
Finally, a {\ffpName} is computed as follows:
\begin{align}
\displaystyle
\fpmap^{fuse} = 0.5*\fpmap + 0.25*\fcmapCeil + 0.25*\fcmapFloor.
\end{align}
\fref{fig:floorplan_fitting} (a) illustrates the above process.
The probability map $\fpmap^{fuse}$ is binarized using a threshold of $0.5$. A bounding rectangle of the largest connected component is computed for later use.
Next, we convert the binary image to a densely sampled piece-wise linear closed loop and simplify it using the Douglas-Peucker algorithm (see~\fref{fig:floorplan_fitting} (b)).
We run a regression analysis on the edges and cluster them into sets of axis-aligned horizontal and vertical lines. 
These lines divide the bounding rectangle into several disjoint grid cells (see~\fref{fig:floorplan_fitting} (c)).
We define the shape of the 2D floor plan as the union of grid cells where the ratio of floor plan area is greater than $0.5$ (see~\fref{fig:floorplan_fitting} (d)).
%
%

\if 1
\begin{itemize}
    \item combine the predicted floor plan maps
    \item thresholding the map by 0.5, finding the bounding rectangle.
    \item find approx polygon (Douglas-Peucker algorithm, using opencv2.approxPolyDP implementation).
    \item regression analysis for each polygon edge. classifier each to the horizon or vertical line by the slope.
    \item clustering lines if distance < $5\%$ diagonal length both horizon and vertical lines. (middle)
    \item overlap with FP map(left) and full up the block which confidence is >= 0.5 (right)
\end{itemize}
\fi

%% file: figures/PL_fig_floorplan_fitting.tex

\begin{figure}[!t]
    \centering
    \includegraphics[width=\columnwidth]{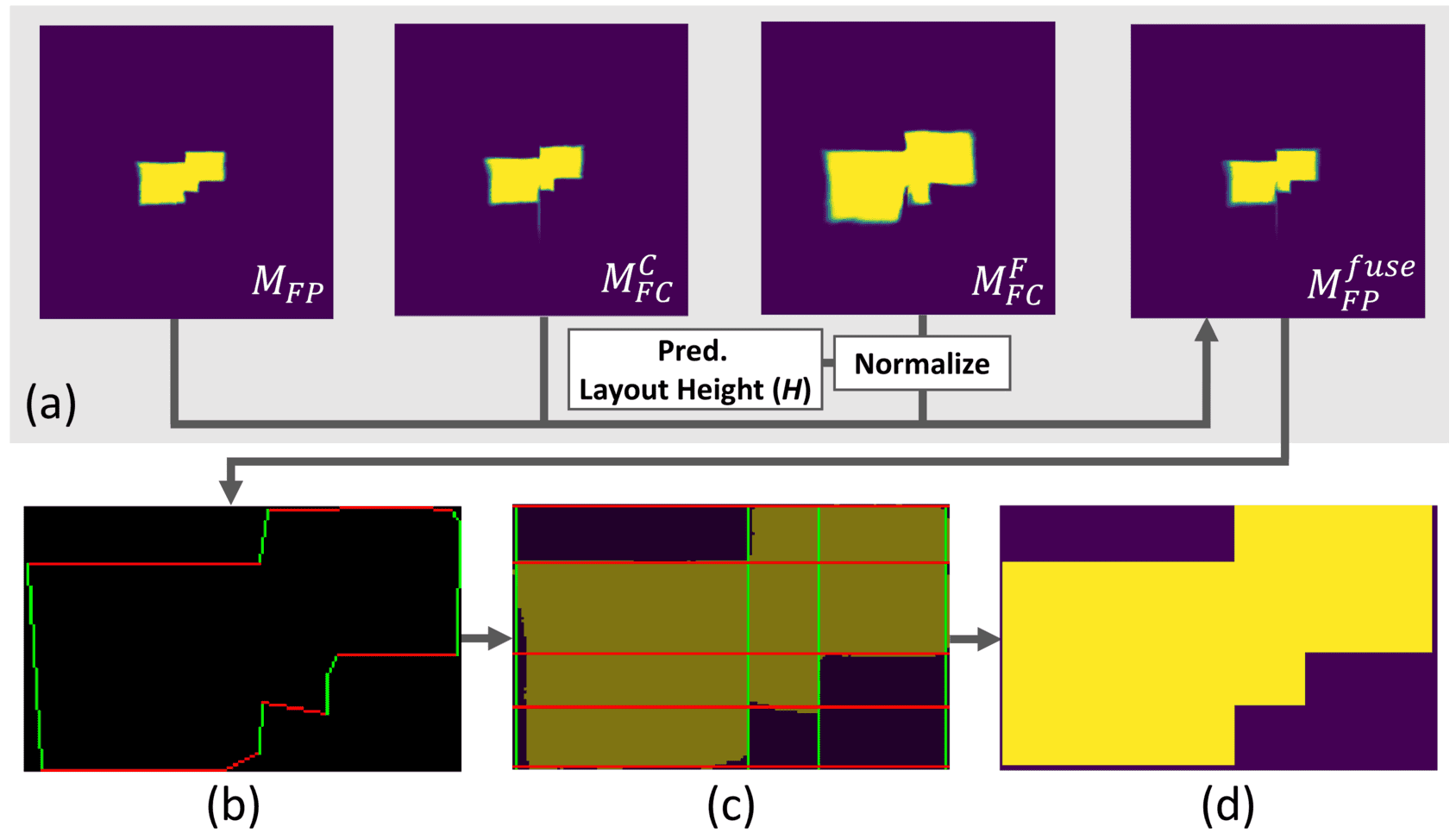}
    \caption{
    \tb{2D floor plan fitting.}
    (a) The probability maps that our network outputs are fused to a {\fpName} $\fpmap^{fuse}$. 
    (b) We apply image thresholding to $\fpmap^{fuse}$ and fit a polygon shape to the floor plan region.
    (c) The polygon edges are regressed and clustered into two sets of horizontal lines (red) and vertical lines (green).
    (d) The final floor plane shape is defined by grids in (c) where the ratio of floor plan area is greater than $0.5$.
    }
    \label{fig:floorplan_fitting}
\end{figure}

%% file: PL_Dataset.tex

\input{./figures/PL_fig_dataset.tex}

\section{{\dsName} dataset}
\label{sec:dataset}
%
%
A dataset that contains a sufficient number of 3D room layouts with different numbers of corners is crucial for training as well as testing our network.
Unfortunately, existing public domain datasets, such as the PanoContext~\cite{panocontext} dataset and the Stanford 2D-3D dataset labeled by Zou~\etal~\cite{layoutnet}, contain mostly layouts with a simple cuboid shape.
To prove that our framework is flexible enough to deal with rooms with an arbitrary number of corners, we introduce a new dataset, named {\dsName}, that contains over 2500 indoor panoramas and annotated 3D room layouts.
We classify each room according to its layout complexity measured by the number of corners in the floor plan.
\tref{tbl:dataset} shows the statistics of the dataset and a few visual examples can be found in~\fref{fig:dataset}.
%
%
%
The source panoramic images in the {\dsName} dataset are collected from two sources. 
The first one is a subset of the SUN360 dataset~\cite{sun360}, which contains \markred{593} living rooms and bedrooms panoramas.
%
%
%
The other source is a real estate database with \markred{1980} indoor panoramas acquired from a real-estate company.
We annotate the 3D layouts of these indoor panoramas using a custom-made interactive tool as explained below.
%

\input{./figures/PL_tbl_eval_full.tex}
\input{./figures/PL_tbl_eval_4corners.tex}

\paragraph{Annotation tool.}

To annotate the 2D indoor panoramas with high-quality 3D room layouts, we developed an interactive tool to facilitate the labeling process. The tool first leverages existing automatic methods to extract a depth map~\cite{fcndepth} and line segments~\cite{panocontext} from the input panorama. Then, an initial 3D Manhattan-world layout is created by sampling the depth along the horizontal line in the middle of the panorama.
The tool allows the users to refine the initial 3D layout through a set of intuitive operations, including (i) pushing/pulling a wall; (ii) merging multiple walls; and (iii) splitting a wall. 
It also offers a handy function to snap the layout edges to the estimated line segments during the interactive editing to improve the accuracy.
%
%
%
%
%
%

\if 1
\fuen{
To prove our proposed method can be applied to complex scenario, we develop our own annotator and a new dataset Complex360 is collected. Our dataset is the first dataset which not only provides complex indoor panoramas along with corresponding labels and also has the largest numbers of labeled panoramas in the world. We will release our source code and the dataset along with annotator once this paper is published.
}

\fuen{
Our Complex360 mainly consists of two parts, SUN360 and our own data. For SUN360, we re-label the panoramas used in PanoContext\cite{panocontext} dataset. The reason why we re-label it is that there are some original labels seems trying to approximate a L-type room as a cuboid one. For our own data, we collect about 3000 indoor panoramas from the web service of our cooperative company and we remove the panoramas which is non-mahhantan or confusing for labeling. The overall numbers of different kinds of layout is shown in Tab.~\ref{tab:dataset}. Our full dataset Complex360 is the merge of SUN360 and our own data. We randomly split 2170 for training and 404 for testing. \textbf{TODO: add some dataset example}
}

\subsection{Annotator}
\label{sec:annotator}
\fuen{
Our annotator takes one panorama, which camera level has been already aligned, as input. First, we will generate an initial layout by a simple depth estimator pretrained in Matterport3D\cite{matterport3D}. We can simply fit a manhattan layout by fitting the gradient direction along the x-axis of panoramas. To make this tool much user friendly, we define 4 simple operations which are \textbf{push}, \textbf{pull}, \textbf{split} and \textbf{merge}. These operations can be applied to all walls, ceiling and floor while we can still fit a corresponding manhattan layout in real-time. After finishing labeling, we will generate the corresponding ground truth of $M_{FC}$, $M_{FP}$, $H$ and the 3D layout.
}
\fi

%% file: figures/PL_fig_dataset.tex

\begin{table}[!t]
    \centering
    \caption{Statistics of the {\dsName} dataset.}
    \begin{tabular}{|c|c|c|c|c|}
    \hline
    4 corners & 6 corners & 8 corners & 10+ corners & Total \bigstrut\\
    \hline
    1246  & 950  & 316   & 61    & 2573 \bigstrut\\
    \hline
    \end{tabular}%
\label{tbl:dataset}
\end{table}

\begin{figure}[h]
    \centering
    \includegraphics[width=\columnwidth]{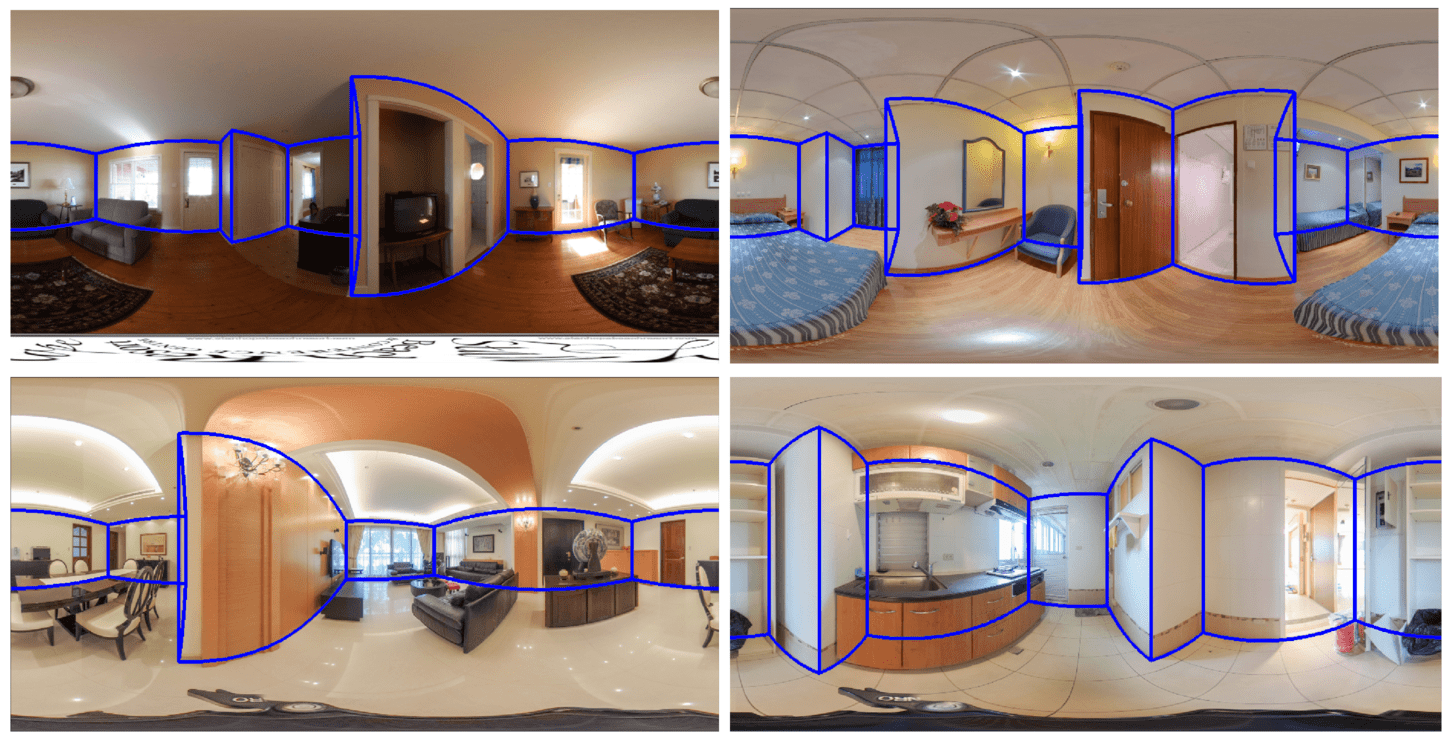}
    \caption{Few example panoramas in {\dsName}. The annotated 3D room layouts are drawn as blue wireframes.}
    \label{fig:dataset}
\end{figure}

%% file: figures/PL_tbl_eval_full.tex

\begin{table*}[!t]
\centering
\resizebox{\textwidth}{!}{
    \begin{tabular}{|c|c|c||c|c||c|c||c|c||c|c|}
    \hline
    \multirow{2}[4]{*}{Method} & \multicolumn{2}{c||}{Average} & \multicolumn{2}{c||}{4 corners} & \multicolumn{2}{c||}{6 corners} & \multicolumn{2}{c||}{8 corners} & \multicolumn{2}{c|}{10+ corners} \bigstrut\\
\cline{2-11}  & \metricND{2}    & \metricND{3}    & \metricND{2}    & \metricND{3}    & \metricND{2}    & \metricND{3}    & \metricND{2}    & \metricND{3}    & \metricND{2}    & \metricND{3} \bigstrut\\
    \hline
    LayoutNet~\cite{layoutnet} & 65.84 & 62.77 & 80.41 & 76.6  & 60.5  & 57.87 & 41.16 & 38.61 & 22.35 & 21.52 \bigstrut[t]\\
    ours (fc-only) & 75.2  & 72.02 & 76.75 & 73.27 & 76.04 & 73.06 & 70.8  & 67.89 & 56.42 & 54.2 \\
    ours (fp-only) & 75.75 & 72.18 & 79.66 & 75.54 & 75.42 & 72.23 & 70.51 & 67.39 & 51.03 & 48.57 \\
    ours (w/o fusion) & 78.52 & 74.8  & 81.77 & 77.57 & 78.5  & 75.1  & 73.61 & 70.37 & 57.01 & 54.12 \\
    ours (full) & \textbf{80.53} & \textbf{77.2} & \textbf{82.63} & \textbf{78.91} & \textbf{80.72} & \textbf{77.79} & \textbf{78.12} & \textbf{74.86} & \textbf{63.1} & \textbf{59.72} \bigstrut[b]\\
    \hline
    \end{tabular}}%
\caption{\tb{Quantitative evaluation on the {\dsName} dataset.} We compare our method with the LayoutNet~\cite{layoutnet}, and conduct an ablation study using different configurations of our method. Bold numbers indicate the best performance.}
\label{tbl:eval_full}
\end{table*}

%% file: figures/PL_tbl_eval_4corners.tex

\begin{table*}[!t]
\centering
\resizebox{\textwidth}{!}{
    \begin{tabular}{|c|c|c||c|c||c|c||c|c||c|c|}
    \hline
    \multirow{2}[4]{*}{Method} & \multicolumn{2}{c||}{Average} & \multicolumn{2}{c||}{4 corners} & \multicolumn{2}{c||}{6 corners} & \multicolumn{2}{c||}{8 corners} & \multicolumn{2}{c|}{10+ corners} \bigstrut\\
\cline{2-11}          & \metricND{2}    & \metricND{3}    & \metricND{2}    & \metricND{3}    & \metricND{2}    & \metricND{3}    & \metricND{2}    & \metricND{3}    & \metricND{2}    & \metricND{3} \bigstrut\\
    \hline
    LayoutNet~\cite{layoutnet} & 71.31 & 67.91 & 80.69 & 76.82 & 68.95 & 65.83 & 50.31 & 47.23 & 44.53 & 42.51 \bigstrut[t]\\
    Ours (full) & \textbf{77.87} & \textbf{74.16} & \textbf{82.42} & \textbf{78.3} & \textbf{77.19} & \textbf{73.74} & \textbf{70.81} & \textbf{67.55} & \textbf{54.05} & \textbf{50.96} \bigstrut[b]\\
    \hline
    \end{tabular}}%
\caption{\tb{Quantitative evaluation on the subset of {\dsName} dataset.} We compare with LayoutNet~\cite{layoutnet} using a training set that contains only rooms with cuboid layout (4 corners). Bold numbers indicate the best performance.}
\label{tbl:eval_4corners}
\end{table*}

%% file: PL_Experiments.tex

\input{./figures/PL_fig_results.tex}

\section{Experiments}
\label{sec:exp}

We compare our method to LayoutNet~\cite{layoutnet}, a state-of-the-art method in room layout estimation, through a series of quantitative and qualitative experiments on our {\dsName} dataset and the PanoContext~\cite{panocontext} dataset. We also conduct ablation study with several alternative configurations of our method.
%
We adopt 2D and 3D Intersection over Union (IoU) to evaluate the accuracy of the estimated 2D floor plans and 3D layouts, which is a standard metric in similar tasks~\cite{deeplabv3plus2018}.
%
All the experiments used the same hyper-parameter discussed in~\sref{sec:training}.
\fref{fig:visual} shows a few 3D room layouts with different numbers of corners estimated using our method.
Please refer to the supplementary materials for more results in the following experiments.

\input{./figures/PL_fig_comparison.tex}

\paragraph{Evaluation on the {\dsName} dataset.}
%
To train both LayoutNet~\cite{layoutnet} and our {\networkName} on the {\dsName} dataset, we randomly selected \markred{2169} panoramas for training and took the remaining \markred{404} panoramas for testing.
We further classify the testing panoramas according to their numbers of corners.
We run LayoutNet using the codes and default hyper-parameter released by the authors.
The quantitative comparison with LayoutNet is shown in~\tref{tbl:eval_full}.
We observe that LayoutNet delivers good performance on cuboid-shaped rooms (4 corners), similar to the numbers reported in their paper. However, the accuracy drops significantly as the number of corners increases.
In comparison, Our {\networkName} not only outperforms LayoutNet on cuboid-shaped rooms by a small margin (around $2\%$), but also performs well on rooms with larger numbers of corners. This leads to an overall performance gain of $\sim14\%$ in both 2D and 3D metrics when compared to LayoutNet.

Since the 3D layout optimization and the hyper-parameter of LayoutNet were tuned on a dataset that contains mostly cuboid-shaped rooms, we conducted another experiment by training both networks on a revised training set that excludes rooms of non-cuboid layouts, while keeping the testing set untouched.
%
\tref{tbl:eval_4corners} shows the quantitative results.
Note that while the performance of LayoutNet improves, our method still outperforms on all kinds of rooms.

From the qualitative comparison shown in~\fref{fig:comparison}, we can observe a strong tendency of LayoutNet to predict the rooms to be cuboid-shaped, possibly due to the constraints imposed in their 3D layout optimization.
In comparison, our method simplifies the problem by directly predicting a Manhattan-world floor plan without any assumptions about the numbers of corners. We conjecture that this is a main reason why our method outperforms LayoutNet, especially with rooms with more than four corners.

%
We also conducted an ablation study that evaluates the performance of our method in different configurations as follows:
%
1) \emph{ours(fc-only)}: only panorama-view branch,
2) \emph{ours(fp-only)}: only ceiling-view branch, and
3) \emph{ours(w/o fusion)}: our full model but without feature fusion.
The quantitative results in~\tref{tbl:eval_full} shows that jointly training both branches leads to better performance than training only one of them. In addition, adding feature fusion between the two branches further improves the performance.


%

\input{./figures/PL_tbl_eval_panocontext.tex}

\paragraph{Evaluation on the PanoContext and Stanford 2D/3D datasets.}
LayoutNet provided quantitative results on the PanoContext~\cite{panocontext} dataset with \markred{414} panoramas for training and \markred{53} panoramas for testing. All rooms are labeled as cuboid-shape. To compare, we trained our network on the same dataset. The quantitative comparison is shown in~\tref{tbl:eval_panocntxt}. Our model outperforms LayoutNet by a small margin.

We also evaluate our model on the Stanford 2D-3D~\cite{stfd2d3d} dataset with annotations labeled by LayoutNet~\cite{layoutnet}. The dataset includes 404 panoramas for training and 113 panoramas for testing. The last column in~\tref{tbl:eval_panocntxt} shows the quantitative result on the Stanford 2D-3D~\cite{stfd2d3d} dataset.


%
%
%

\paragraph{Timing.}
An end-to-end computation takes three main steps - 1) an alignment process to align the input panorama with a global coordinate system, 2) {\fpName} prediction by our neural network, and 3) 2D floor plan fitting. Step 1) is most time-consuming, which takes about 13.37s measured on a machine with a single NVIDIA 1080ti GPU and Intel i7-7700 3.6GHZ CPU. Step 2) takes only 34.68ms and step 3) takes only 21.71ms. 

Compared to LayoutNet, they carry out the same alignment process and their neural network prediction is also very fast (39ms). However, they needed another very time-consuming 3D layout optimization step in the end, which takes 30.5s. In summary, an end-to-end computation by LayoutNet takes about 43.9s while our method takes about 13.4s, a speed up of 3.28X.

\if 1
\fuen{
In this section, we will compare the result of our proposed method with the one of other baselines. In general, we conduct our experiment on both original PanoContext dataset and our proposed dataset Complex360.
}

\fuen{
We compare our method with current state-of-the-art method \textbf{LayoutNet}\cite{layoutnet}. In addition, we have done several ablation study for our method. As our proposed dataset consists of several kinds of layout, which are 4, 6, 8 and 10+ (more than 10) corners, we show our quantitative result for each of them. We use both 2D and 3D IoU as our evaluation metrics. 

Tab.~\ref{tab:result-complex360} shows our accuracy in our Complex360. \textbf{Ours (fp-only)} and \textbf{Ours (fc-only)} are our proposed method without using feature fusion technique and only use one of the outputs from \ceilingbranch and \panoramabranch for post-process, respectively. \textbf{Ours (w/o fusion)} is our proposed method without using feature fusion technique but use both outputs from \ceilingbranch and \panoramabranch for for post-process. From Tab.~\ref{tab:result-complex360}, our proposed method \textbf{Ours (full)} outperforms other baseline.

Although \textbf{LayoutNet} can infer accurate layout for 4 corner case, its accuracy start to decrease significantly when the number of corners is larger than 6 because it is designed only for cuboid (4 corners) and L-type (6 corners) layout. Compared with \textbf{Ours (fp-only)} and \textbf{Ours (fc-only)}, we can prove our two-branch architecture is necessary because using only one of them cannot provide enough information for layout prediction. However, although \textbf{Ours (w/o fusion)} uses both two branches, the lack of both local and global information makes its two sub-networks unable to fit the layout very well. As a result, we prove that our proposed method have the capability of predicting 360$^\circ$ layout for both simple and complex scenario.
}

\fuen{
To further compare our method with \textbf{LayoutNet}, we conduct our experiments on both original PanoContext\cite{panocontext} dataset, which is a cuboid (4 corners) dataset used by LayoutNet, and the subset of our Complex360 consists of only cuboid (4 corners) layout. Tab.~\ref{tab:result-panocontext} shows the quantitative result in PanoContext dataset. From this result, we prove our method can also outperform LayoutNet in simple scenario such as cuboid layout. In Tab.~\ref{tab:result-complex360-4corners}, we show the quantitative result on the subset of Complex360. The network is trained on only 4 corners but we also evaluate it on other complex rooms which have more than 4 corners. From this result, although our training data only consists of 4 corners layout, our method can still infer a reasonable cuboid layout to approximate other complex cases (4, 6 and 8 corners) and still achieve higher 2D/3D IoU . 
 }
 

 \fi

%% file: figures/PL_fig_results.tex

 \begin{figure*}[!t]
  \centering
  \includegraphics[width=0.98\textwidth]{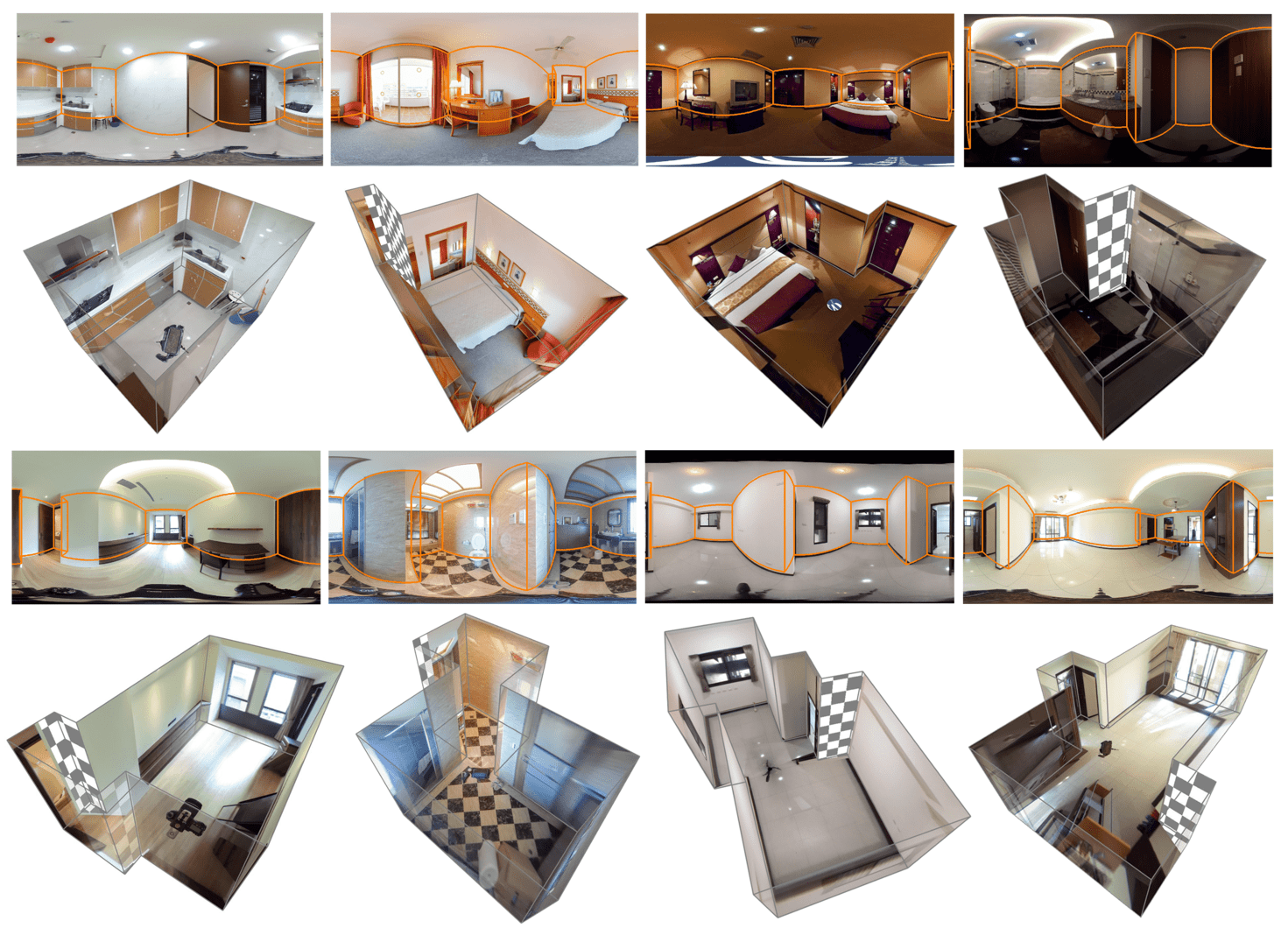}
  \caption{
  \tb{Visual results.} Given a single RGB panorama, our method automatically estimates the corresponding 3D room layout. Our method is flexible to handle more complex room layout beyond the simple cuboid room. The checkerboard patterns on the walls indicate the missing textures due to occlusion.
  }
  \label{fig:visual}
\end{figure*}

%% file: figures/PL_fig_comparison.tex

\begin{figure*}[!t]
  \centering
  \includegraphics[width=\textwidth]{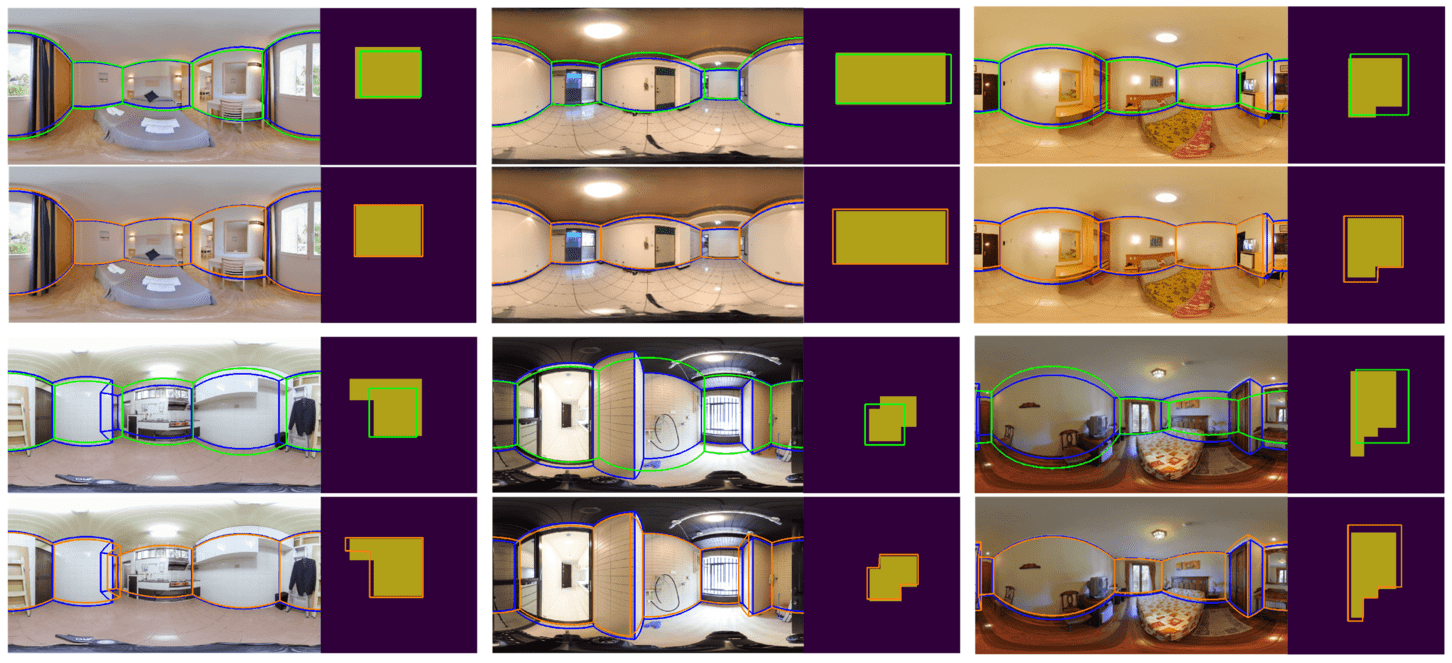}
  \caption{
  \tb{Qualitative comparison with LayoutNet~\cite{layoutnet}.}
  The 3D room layouts generated by LayoutNet\cite{layoutnet} (green lines) and our method (orange lines). Results are displayed on both the equirectangular {\panoView} (\emph{left}) and floor plan view (\emph{right}), where the blue lines and yellow solid shapes represent the ground truth, respectively.    
  }
  \label{fig:comparison} 
\end{figure*}

%% file: figures/PL_tbl_eval_panocontext.tex

\begin{table}[!b]
\centering
\caption{\tb{Quantitative evaluation on the PanoContext~\cite{panocontext} and Stanford 2D/3D~\cite{stfd2d3d} datasets in \metricND{3}}}
\resizebox{1.0\columnwidth}{!}{
    \begin{tabular}{|c|c|c|}
    \hline
    Method & PanoContext & Stanford 2D-3D \bigstrut\\
    \hline
    \multirow{2}[2]{*}
         ~LayoutNet~\cite{layoutnet} & 74.48 \bigstrut[t] & 76.33 \bigstrut[t] \\
         ~Ours (full) & \textbf{77.42} \bigstrut[b] & \textbf{79.36} \bigstrut[b] \\
    \hline
    \end{tabular}}
\label{tbl:eval_panocntxt}
\end{table}

%% file: PL_Conclusion.tex

\input{./figures/PL_fig_limitations.tex}

\section{Conclusion}
We present an end-to-end deep learning framework, called {\networkName}, for estimating 3D room layouts from a single RGB panorama.
We propose a new network architecture that consists of two encoder-decoder branches for analyzing features from two distinct views of the input panoramas, namely the equirectangular {\panoView} and the perspective {\ceilView}.
The two branches are connected through a novel feature fusion scheme and jointly trained to achieve the best accuracy in the prediction of 2D floor plan and layout height.
To learn from complex layouts, we introduce a new dataset, {\dsName}, which contains {\dsSize} indoor panoramas of Manhattan-world room layouts with various complexity.  
Both the quantitative and qualitative results demonstrate that our method outperforms the current state-of-the-art in prediction accuracy, especially with rooms with more than four corners, and take much less time to compute the final 3D room layouts.

\paragraph{Limitations and future work.}
Our method has the following limitations:
i) without knowing the object semantics, our network might get confused with the rooms that contains mirrors or large occluding objects as shown in~\fref{fig:limitation}; and
ii) our approach of 3D layout estimation involves heuristics and assumptions that might over- or under-estimate the underlying {\fpName} and also restrain the results to Manhattan world.
We propose to explore the following directions in the near future. 
First, introducing the object semantics, i.e., segmentation and labels, to the network architecture could potentially improve the accuracy by ignoring those distracting and occluding objects from the floor plan prediction.
Second, designing a principled algorithm for a more robust 3D layout estimation, e.g., no Manhattan-world assumption and support rooms with curve shapes. 
Last but not the least, we believe that even better results can be achieved by experimenting with a larger range of encoders for our network architecture.

\paragraph{Acknowledgements.} The project was funded in part by the KAUST Office of Sponsored Research (OSR) under Award No. URF/1/3426-01-01, and the Ministry of Science and Technology of Taiwan (107-2218-E-007-047- and 107-2221-E-007-088-MY3).


%

%% file: figures/PL_fig_limitations.tex

\begin{figure}[!b]
\begin{center}
  \includegraphics[width=\columnwidth]{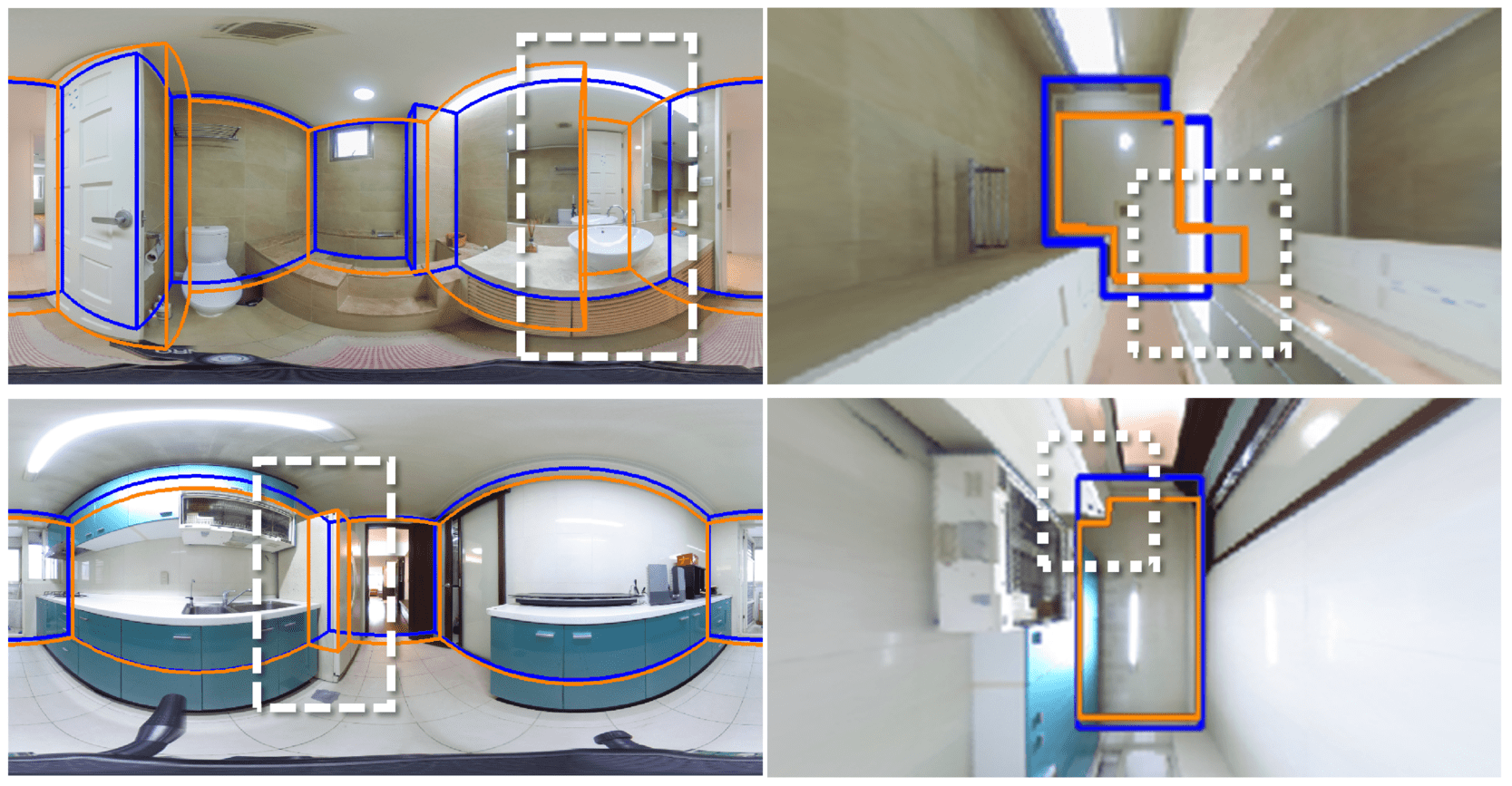}
  \caption{\tb{Limitations.} 
  Two failure cases generated by our method (orange lines) due to the lack of object semantics. (\emph{Top}) Our method is misled by the reflection of mirror. (\emph{Bottom}) The boundary of floor plan is occluded by the refrigerator. The ground truth layout is rendered in blue.}
  \label{fig:limitation}
\end{center}
\end{figure}

%% file: PL_Supplementary.tex

\section{Comparisons to LayoutNet~\cite{layoutnet}}
\label{sec:supplementary}

\subsection{Re-training LayoutNet}

For the comparisons with LayoutNet, we used the neural network and the 3D optimization module provided by the authors of LayoutNet.
According to our analysis, the part of the code that concerns the 3D optimization module is specially designed for cuboid-shape rooms and works very well.
However, the part designed for non-cuboid shaped rooms has multiple issues and is not sufficiently robust.
Our analysis indicates that the algorithm has the best performance when it only tries to fit rooms with 4 corners.
To provide the best possible version of LayoutNet, we experimented with three different settings in network training and report the results here.
1) LayoutNet with the original weights pre-trained by the authors.
2) LayoutNet retrained with the subset of our {\dsName} dataset, which contains only cuboid layouts (4 corners).
3) LayoutNet retrained with our complete {\dsName} training dataset.
The performance comparison is shown in~\tref{tbl:eval_layoutnet}.
We can observe that LayoutNet's performance improved significantly when it is re-trained on our {\dsName} dataset.
It is further improved when it is re-trained with cuboid-shaped rooms only since the 3D optimization module works correctly only for such rooms.
In the paper, we therefore reported results for the best version of LayoutNet that we could create, i.e. LayoutNet re-trained on our {\dsName} dataset with rooms of cuboid layouts (4-corners) only, followed by their 3D optimization module for cuboid-shaped rooms.


\begin{table}[h]
\centering
\caption{\tb{Quantitative evaluation of LayoutNet on different training sets.}}
\resizebox{0.7\columnwidth}{!}{
    \begin{tabular}{|c|c|c|c|}
    \hline
    Metric & Training set & Average \bigstrut\\
    \hline
    \multirow{2}[3]{*}{3D IoU (\%)} & pre-trained & 57.96 \bigstrut[t]\\
          & Realtor360 (4-only) & \textbf{67.91} \bigstrut[b]\\
         & Realtor360 (all) & 62.77 \bigstrut[b]\\
    \hline
    \end{tabular}}
\label{tbl:eval_layoutnet}
\end{table}

\subsection{More Qualitative Comparisons}
\fref{fig:comparison1} and \fref{fig:comparison2} show more results of qualitative comparisons with LayoutNet.

\if 1
Note that the cuboid-shape layout optimization presented by LayoutNet may still generate visible artifacts when applied to non-cuboid shaped rooms as shown in \fref{fig:comparison3} (2nd col., 9th row), \fref{fig:comparison7} (4th col., 7th row), \fref{fig:comparison9} (2nd col., 7th row), and \fref{fig:comparison16} (2nd col., 5th row).
\fi

%

\section{More Visual Results of {\networkName}}
As the complexity of the estimated room layout is best viewed in 3D, we show more 3D room layouts with different numbers of corners generated by our method in~\fref{fig:visual1} and \fref{fig:visual2}.

\newcommand{\compareCaption}{\tb{Qualitative comparison with LayoutNet~\cite{layoutnet}.} The esitmated 3D room layouts in equirectangular panorama-view. (\textit{Blue lines}) Ground truth. (\textit{Green lines}) LayoutNet's result. (\textit{Orange lines}) Our result.}

\begin{figure*}[t!]
  \centering
  \includegraphics[width=0.975\textwidth]{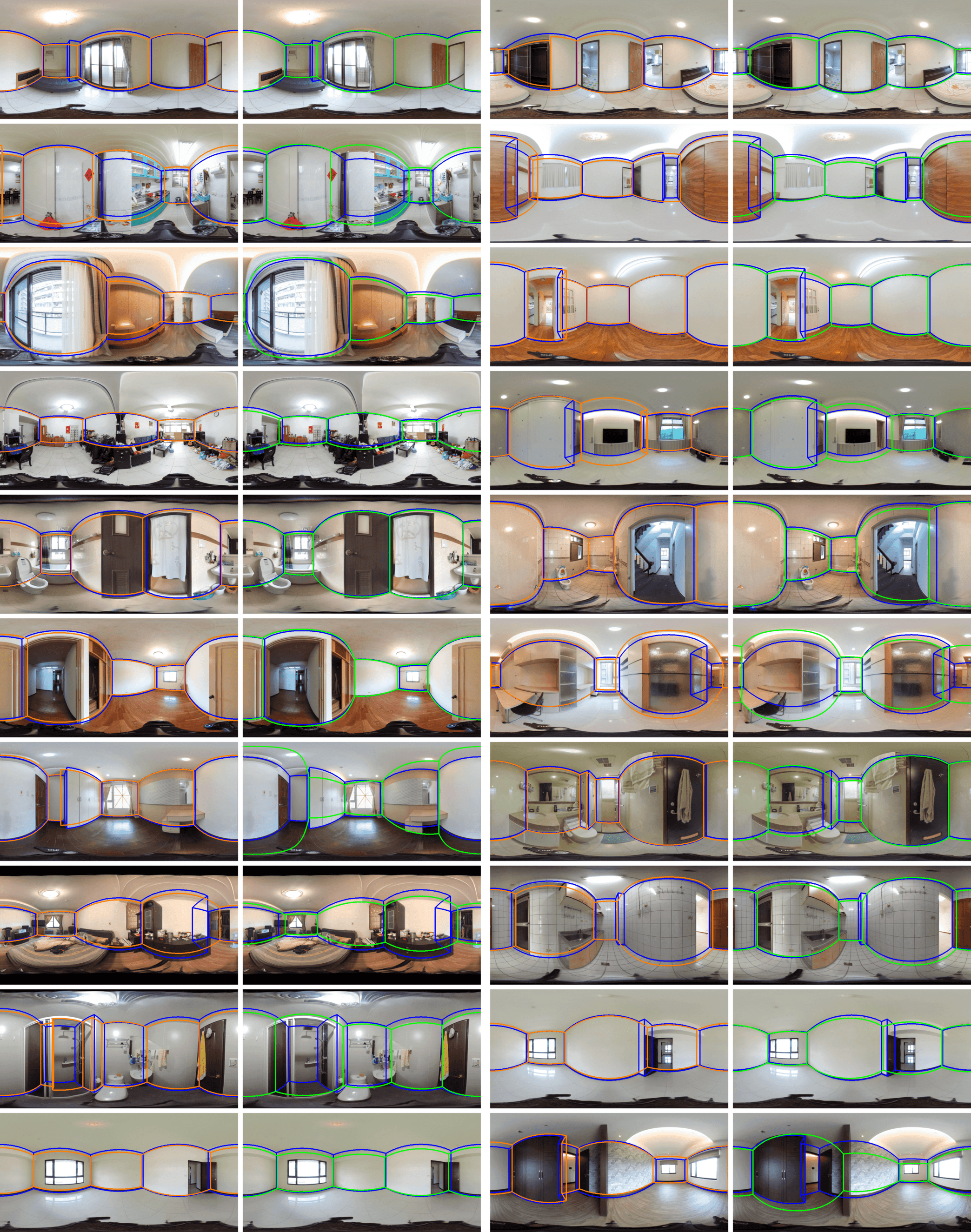}
  \caption{\compareCaption}
  \label{fig:comparison1}
\end{figure*}

\begin{figure*}[t!]
  \centering
  \includegraphics[width=0.975\textwidth]{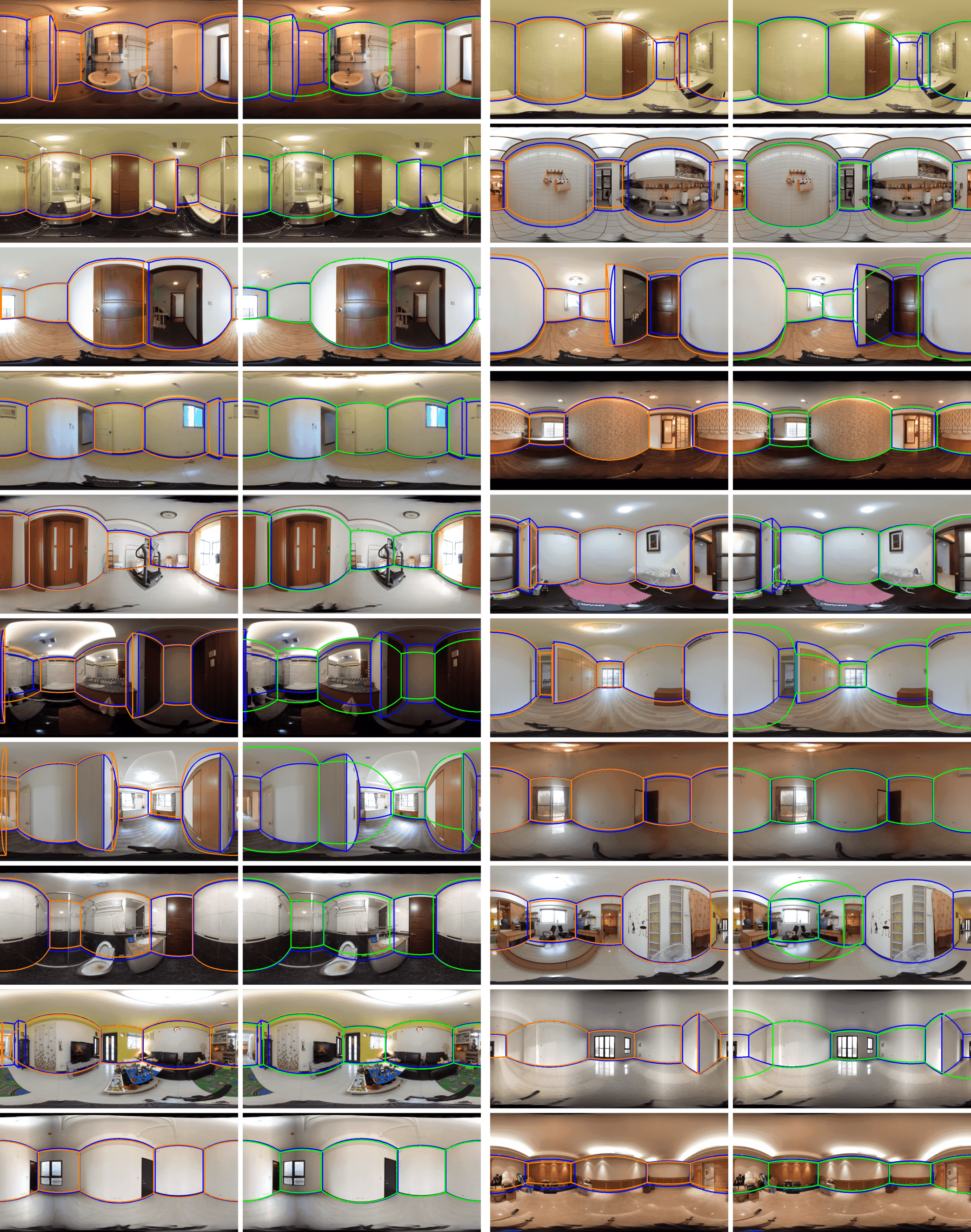}
  \caption{\compareCaption}
  \label{fig:comparison2}
\end{figure*}

\newcommand{\visualCaption}{\tb{Visual results.} The 3D room layouts estimated by~{\networkName} are displayed in equirectangular {\panoView} (\textit{top}) and 3D rendering with texture mapping (\textit{bottom}). Note that the checkerboard patterns on the walls indicate the missing textures due to occlusion.}

\begin{figure*}[t!]
  \centering
  \includegraphics[width=0.875\textwidth]{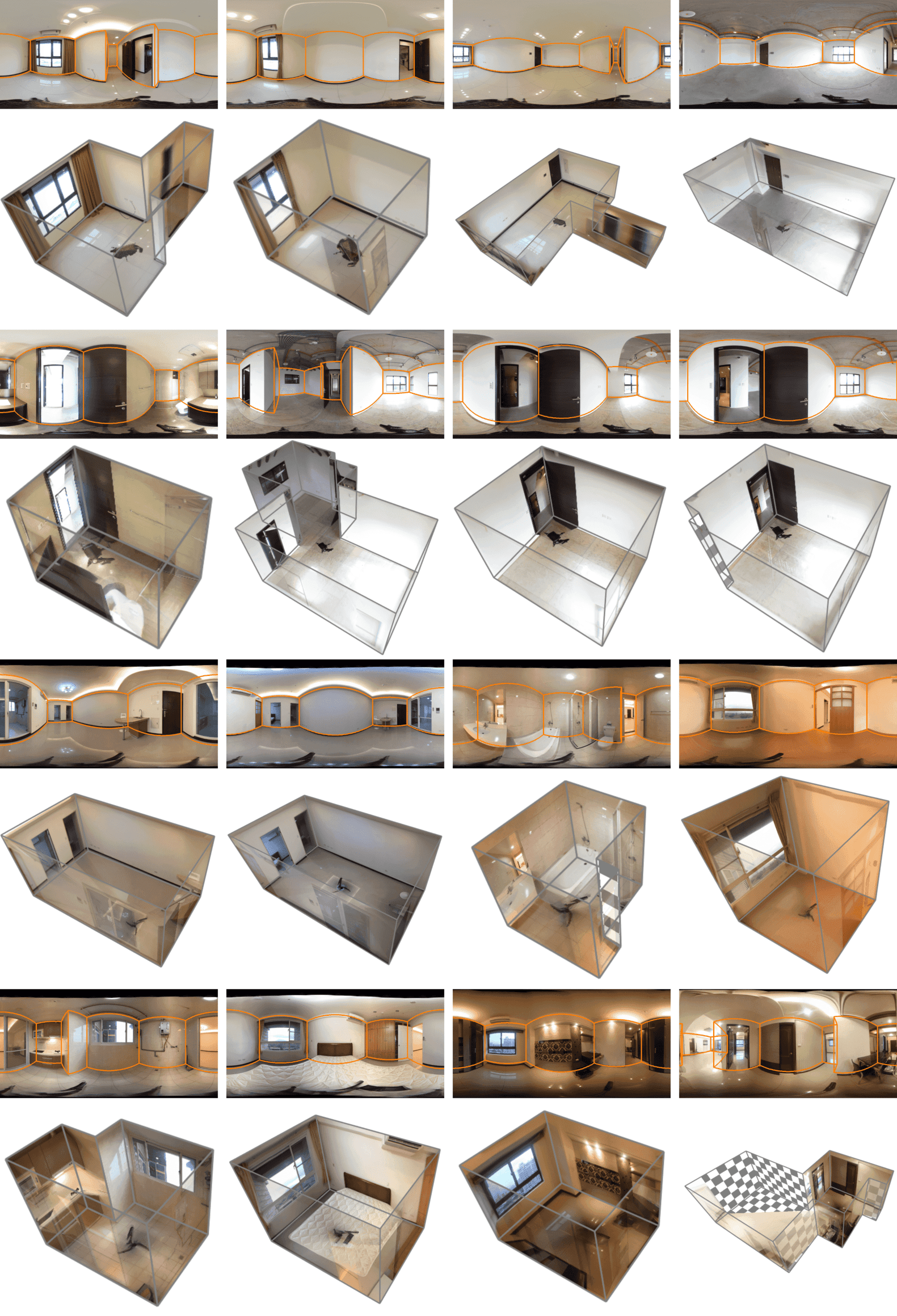}
  \caption{\visualCaption}
  \label{fig:visual1}
\end{figure*}

\begin{figure*}[t!]
  \centering
  \includegraphics[width=0.875\textwidth]{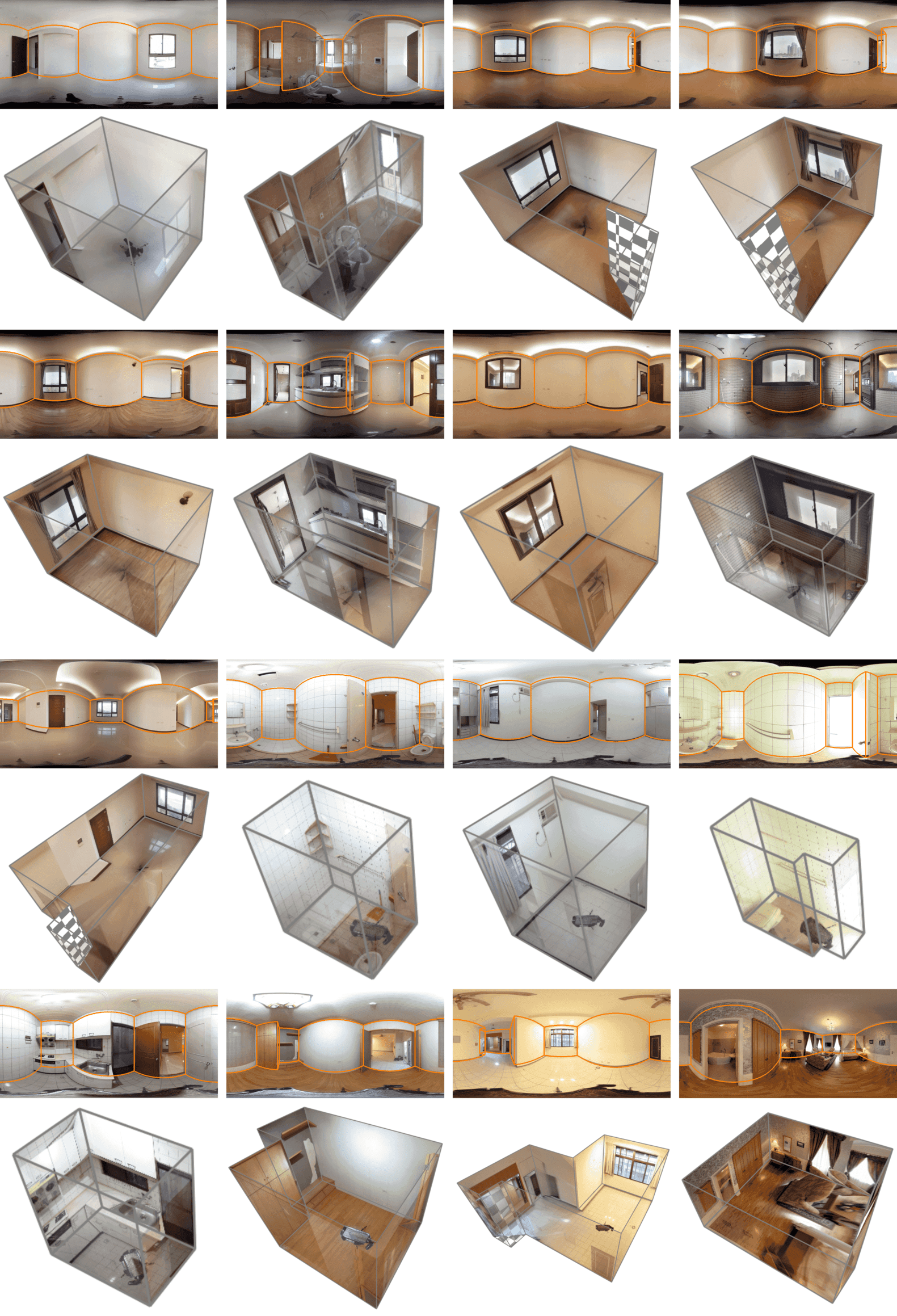}
  \caption{\visualCaption}
  \label{fig:visual2}
\end{figure*}

\if 1
{\color{red}
\paragraph{Corner prediction performances of LayoutNet.} [TO REMOVE?] In~\fref{fig:comparison}, we show the corner prediction results of LayoutNet on our dataset (a cuboid room) using their pre-trained model, a model re-trained on our dataset with cuboid-shaped rooms only, and re-trained on our dataset with all kinds of rooms. We can see that the performance improved significantly after re-training. We also observed that the corner prediction network of LayoutNet may ...
}

\begin{figure}[h]
  \centering
  \includegraphics[width=\columnwidth]{sec1}
  \caption{
  {\color{red}
  [TO REMOVE?] \tb{Comparison example} The layout reuslt(left column), corner map(middle column) and boundary map (right column) output by LayoutNet. Form top to bottom row are official pre-train model, training on Realtor360 (4-only) and training on Realtor360 (all).
  }
  }
  \label{fig:comparison}
\end{figure}

\subsection{LayoutNet failed case(?}
While we were running the LayoutNet, there is some obvious fail case.  Since the final 3D layout optimization is base on the corner and boundary prediction(especially corners), the method is sensitive to the quality of the corner map.  If the map has noise, the method would be easy to be affected and generate the layout far from the cuboid shape. This issue is easy to be observed while the input is non-cuboid or the scene has ambiguous structure likes decoration or open door.
\fi